\documentclass[review]{elsarticle}

\usepackage{lineno,hyperref}
\modulolinenumbers[5]

\journal{Journal Name}

\usepackage{graphicx}
\usepackage{array}
\usepackage{caption}
\usepackage{subcaption}
\usepackage{enumitem}
\usepackage{xspace}
\usepackage{xparse}
\usepackage{amsmath,amssymb}
\usepackage{comment}

\newcolumntype{P}[1]{>{\centering\arraybackslash}p{#1}}
\newcolumntype{M}[1]{>{\centering\arraybackslash}m{#1}}

\graphicspath{ {./img/} }

\usepackage{multirow}
\usepackage{mathtools}

\usepackage{color,soul} 
\definecolor{tangelo}{rgb}{0.80, 0.2, 0.0}
\definecolor{black}{rgb}{0.0, 0.0, 0.0}

\newcommand{\generator}{G} 
\newcommand{\discriminator}{D}
\newcommand{\real}{x}
\newcommand{\latent}{z}
\newcommand{\fake}{G(\latent)}
\newcommand{\E}{\mathbb{E}}

\begin{document}

\begin{frontmatter}
%
%
\title{Generative Adversarial Networks via a Composite Annealing of Noise and Diffusion}
%
%
%
\author{Kensuke Nakamura}
\address{Computer Science Department, Chung-Ang University, Seoul, Korea}
\author{Simon Korman} 
\address{Department of Computer Science, University of Haifa, Israel} 
\author{Byung-Woo Hong\corref{mycorrespondingauthor}}
\address{Computer Science Department, Chung-Ang University, Seoul, Korea}
%
%
%
\cortext[mycorrespondingauthor]{Corresponding author}
\ead{hong@cau.ac.kr}
%
%
%
%
%
%
\begin{abstract}
Generative adversarial network (GAN) is a framework for generating fake data using a set of real examples. However, GAN is unstable in the training stage. In order to stabilize GANs, the noise injection has been used to enlarge the overlap of the real and fake distributions at the cost of increasing variance. The diffusion (or smoothing) may reduce the intrinsic underlying dimensionality of data but it suppresses the capability of GANs to learn high-frequency information in the training procedure. Based on these observations, we propose a data representation for the GAN training, called noisy scale-space (NSS), that recursively applies the smoothing with a balanced noise to data in order to replace the high-frequency information by random data, leading to a coarse-to-fine training of GANs. We experiment with NSS using DCGAN and StyleGAN2 based on benchmark datasets in which the NSS-based GANs outperforms the state-of-the-arts in most cases.
\end{abstract}
%
%
%
\end{frontmatter}
%
%
%
%
%
%
%
%
%
%
%
%
\section{Introduction} 
Generative adversarial network (GAN)~\cite{goodfellow2014generative} is a machine learning framework to generate realistic fake data. GAN learns the probabilistic distribution of the training (real) data using two adversarial networks: the generator that is trained to create realistic fake data from a random seed called the latent vector, and the discriminator that is dedicated to distinguish the reals against fake data. GAN has been studied extensively in the several past years and currently is an essential tool for a wide variety of applications.
\par
However, the training procedure is prone to numerical unstability in GANs~\cite{Nagarajan17gdGANstable,Mescheder2018which,berard2019closer,kurach2019large,Wang2020LSVGDforGAN}.
Since it is a two-player game between the generator and discriminator~\cite{berard2019closer}, the optimization can fall into a local minima where the discriminator reaches a perfect solution first and the generator cannot be trained anymore. 
This failure of GAN severely limits the quality of fake data, resulting in the mode collapse.
%
The failure of GAN occurs when there is almost no overlap between the real distribution with fake distribution~\cite{arjovsky2017towards}, in particular, in the beginning of training when the discriminator can reject fake data with a high confidence~\cite{goodfellow2014generative}. 
Therefore, GANs require stabilization methods in the optimization process in order to obtain better fake data.
\par
The stabilization methods of GAN are different with those for the feed-forward deep networks, e.g., weight-decay~\cite{krogh1992simple} and the momentum~\cite{sutton1986two}, due to the dynamics of the discriminator with generator.
Since the failure of GAN can arise from the use of KL-divergence in loss, a variety of the discrepancy measures have been studied, e.g., ~\cite{mao2017least,Nowozin16fGAN,cai2020utilizing}, including Wasserstein distance~\cite{arjovsky2017towards,arjovsky2017wasserstein} that provides gradients to the generator even with a small overlap of the two distributions.
Albeit, the Wasserstein loss is often inferior to the original loss in the quality of fake data~\cite{Mescheder2018which,kurach2019large}.
%
%
%
%
Another strategy is to inhibit discriminator training based on the loss. To this aim, the gradient regularization penalizes gradients of the discriminator. However, it depends on the model that varies during the training~\cite{kurach2019large}.
\par
This paper focuses on data-based stabilization of the GAN training that  manipulates only the real and fake data independent of the model architecture.
For instance, the repetition of noise injection to both real and fake data  (namely noise-space) enlarges the overlap between probability distributions~\cite{jenni2019noiseGAN}. However, the noise increases the variance of data inevitably.
The repetition of data smoothing based on Gaussian kernels, or the scale-space, is a general technique in machine learning that suppresses the high frequency features such as textures and details in the image so as to make each data more simple to learn by algorithms. 
In the case of GAN, it is expected that the diffusion makes the real data easy to mimic by the generator.
However, we have found that it limits the learning capability of the generator to learn high-frequency information.
\par
Based on these observations, we propose an algorithm for stabilizing the optimization of GANs based on a \textit{noisy scale-space} (NSS) that continuously removes high-frequency information in image while adding noise.
The proposed noisy scale-space enables a coarse-to-fine training of GANs 
in which we can train a generative model using low-level information in data with noise without the increment of data variance, while keeping the current model capable of learning the high-level information.
We also present a synthetic dataset using the Hadamard bases~\cite{townsend2001walsh} that can visualize the true distributions of real and fake data in order to characterize the drawback of the conventional scale-space in GAN optimization.
Then, we perform experiments with the proposed NSS using DCGAN~\cite{radford2015dcgan}. The experimental results based on the major datasets show that the proposed NSS-GAN outperforms other methods in most cases irrespective of the image generation tasks. Specifically, it is shown that the stabilization effect by our method is not the simple summation of the noise-space with the scale-space but due to the use of their mutually complementary relationship.
We also demonstrate that the proposed NSS can improve the accuracy of StyleGAN2~\cite{karras2020training} for high-resolution images.
\par
We relate our method to prior works in Section~\ref{sec:related_works}.
Then we consider a data-based stabilization of GAN training in Section~\ref{sec:preliminary}, followed by the proposed noisy scale-space in Section~\ref{sec:method}.
The effectiveness of our method is demonstrated experimentally in Section~\ref{sec:results} and we conclude in Section~\ref{sec:conclusion}.
%
%
%
%
%
\section{Related works}   \label{sec:related_works}
\par
\vspace{3pt}
\noindent {\bf Loss-based GAN stabilization:} 
The regularization term in the loss has been studied to restrict the update of discriminator.
In a related context, Wasserstein-GAN (WGAN) uses a weight clipping~\cite{arjovsky2017wasserstein} to guarantee the Lipschitz constraint. WGAN-gp uses a gradient penalty~\cite{Sailmans16improved} that can improve the quality of fakes in practice. The spectral normalization~\cite{Miyato17spectral} is an efficient variant of the gradient-penalty. Dragan~\cite{kodali2017convergence} penalizes the sharp gradient of discriminator to real data.
%
We will use the non-saturating loss~\cite{goodfellow2014generative} based on the KL-divergence that is known to be the best choice in practice. 
The non-saturating loss can be further improved using the gradient regularization~\cite{Mescheder2018which} that penalizes the gradient norm of the discriminator.
It is shown that the penalty based on the gradient-norm of the discriminator is equivalent to adding input noise in GAN using $f$-divergence~\cite{Kevin2017regularization}. 
The drawback of gradient regularization is that it depends on the distribution of fakes determined by the generator which changes during training~\cite{kurach2019large}.
\par
\vspace{3pt}
\noindent {\bf Procedure-based GAN stabilization:} 
The two time-scale update rule~\cite{Heusel17TTUR} is a method of using different annealings for discriminator and generator in order to slow the convergence of the discriminator. 
The simultaneous update of the two networks was studied in~\cite{schafer2019implicit}.
Progressive augmentation of GAN~\cite{Zhang19progressive_aug} extends the label noise~\cite{zhang2016understanding} into the GAN framework to perturb the real and fake labels. The one-sided label smoothing replaces the 0 and 1 target labels for the discriminator with smoothed values, like 0.9 or 0.1~\cite{Sailmans16improved,hazan2017adversarial}. 
%
%
%
%
\par
\vspace{3pt}
\noindent {\bf Data-based GAN stabilization:} 
The proposed method can be categorized in data-based methods that manipulate only data.
Lens-GAN~\cite{sajjadi2018tempered} introduces a filtering network that transforms the real data against the discriminator, and therefore it still depends on the networks.
\par
The multi-resolution training of GAN is a topic studied in, e.g., Progressive-GAN~\cite{karras2017progressive} and MSG-GAN~\cite{Karnewar20MSGGAN}, for generating high-resolution images.
It trains a shallow network first using low-resolution images, and gradually increases both the number of network layers and the resolution of data.
The drawback of the multi-resolution training is that it strongly limits the architecture.
\par
The data augmentation is another recent topic in GAN training~\cite{zhao2020differentiable,karras2020training,tran2021data}
in which a multitude of data transformations, typically consist of spatial transformations (e.g., image rotation, flipping, and cropping) with color transformations (e.g., channel permutation and hue rotation) are combined and applied to both real and fake data in order to enlarge the variation of data.
The early works~\cite{zhao2020differentiable,karras2020training} are oriented to avoid over-fitting of GANs to a small set of training examples while the recent works aim to improve the accuracy as we do.
Also DistAug~\cite{jun2020distribution} has considered a mixture of data transformations in a contrastive training of GANs~\cite{chen2019self}.
However, the mixture of transformations in these studies is a black-box.
In contrast, we present a deep understanding of the noise injection and the image diffusion in GAN optimization and then propose a better use of their mutually complementary relationship.
\par
Our method is closely related to the noise injection and the data smoothing.
On one hand, the noise injection flattens the probability distribution of data.
Thus imposing noise to the real data~\cite{sonderby2016amortised,arjovsky2017towards} or both the real and fake data~\cite{jenni2019noiseGAN} enlarges the overlap of the two distributions.
However, adding high-dimensional noise introduces significant variance in the parameter estimation, slowing down the training and requiring multiple samples for counteraction~\cite{Kevin2017regularization,Zhang19progressive_aug}.
On the other hand, the scale-space decreases the dimensionality of data by removing high-frequency information and makes data easy to learn by algorithms~\cite{witkin1987scale,lindeberg2013scale}.
However, the conventional scale-space has a critical side-effect in GAN optimization.
Different from these baselines, we present a data representation that mitigates the side effects of the noise injection and the data smoothing.
\par
Our algorithm also relates to Ambient-GAN~\cite{bora2018ambientgan} that considers a problem in which incomplete real data containing a Gaussian blurring with additive noise are given. However, Ambient-GAN aims to approximate the original distribution from the incomplete measurements using a conditional network, and created fake images are degenerated by the noise and smoothing.
In contrast, we propose a continuous data representation to train GANs, achieving a better quality of fake data than the baseline method using the complete examples. 
%
%
%
%
%
%

%
%
%
\section{Preliminary} \label{sec:preliminary}
\subsection{Generative adversarial networks}
Let us begin with a technical introduction to the generative adversarial networks (GAN).
Given a set of real data ($\real$), GAN aims to generate new data with similar statistics as the real data. GAN consists of the two networks: the generator ($\generator$) that creates fake data ($\fake$) from a random latent vector ($\latent$), and the discriminator ($\discriminator$) that distinguishes the real data against the fake data.
To this aim, a min-max loss ($F$) is defined~\cite{goodfellow2014generative} as 
\begin{equation} \label{eq:GAN-loss-original}
F_{D,G} \bigl( x,G(z) \bigr) \quad \coloneqq \quad
\E_{x \sim p_\text{data}(x)} \bigl[ \log D(x) \bigr] + 
\E_{z \sim p_\text{z}(z)} \bigl[ \log (  1 - D(G(z)) ) \bigr], 
\end{equation}
%
%
where $p_\text{data}(x)$ is the true distribution of the reals, and $p_\text{z}(z)$ is a random probability distribution.
GANs are optimized using a stochastic gradient descent, e.g., Adam~\cite{kingma2014adam}, in an alternative way such that the generator is updated to minimize $F$ while the discriminator is trained to maximize $F$.
\subsection{Data-based stabilization of GAN training}
We consider a data-based stabilization method, similar to~\cite{zhao2020differentiable,karras2020training,tran2021data}, that projects both the reals and fakes to a data space and feed into the GAN loss as
%
%
\begin{flalign}\label{eq:GAN-loss-filtered}
& F_{D,G} \bigl( \Phi_t(x),\Phi_t(G(z)) \bigr),
\end{flalign}
%
%
where $\Phi_t(x)$ and $\Phi_t(G(z))$ are the projected data of reals and fakes using a function $\Phi$ with the time parameter $t$.
%
Using Eq.(\ref{eq:GAN-loss-filtered}), 
the discriminator is trained to distinguish the projected real against the projected fake while
the generator is trained to create fake data such that the projected fakes $\Phi_t(G(z))$ are similar to the projected reals $\Phi_t(x)$.  
The data-based stabilization effect will be imposed via a discrete representation of data
$\{\Phi_T(y),\Phi_{T-1}(y),...,\Phi_2(y),\Phi_1(y),\Phi_0(y)\}$ 
where $\Phi_t(y)$ with larger $t$ is expected to have larger effect
such that the  data is more easy to learn by GANs, 
and $\Phi_0(y) = y$ denotes the original real or fake data.
To this aim, the time parameter ($t$) should start with a large value $T$ and shrink to zero during the training, and the choice of $\Phi$ determines the stabilization effect.
%
%
%
%
%
%
%
%
%
%

%
%
\def \fw {55pt}
\def \pw {45pt}
\begin{figure} [htb]
\centering
\small
%
\begin{tabular}{m{32pt} m{\pw}m{\pw}m{\pw}m{\pw}m{\pw}m{\pw}}
(a) NS &
\includegraphics[height=\fw]{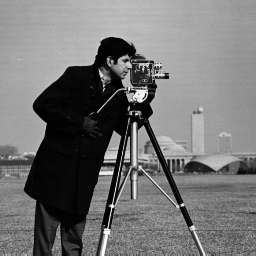} &
\includegraphics[height=\fw]{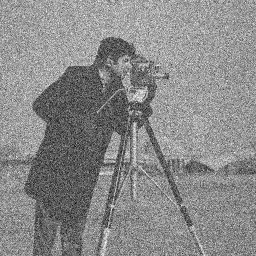} &
\includegraphics[height=\fw]{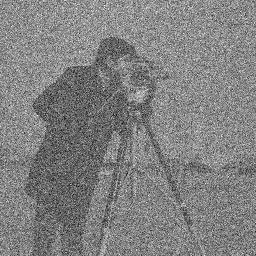} &
\includegraphics[height=\fw]{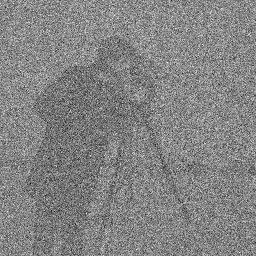} &
\includegraphics[height=\fw]{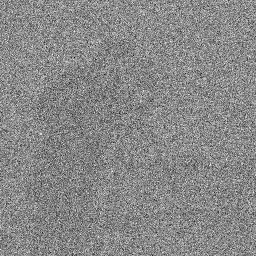} \\
(b) SS & 
\includegraphics[height=\fw]{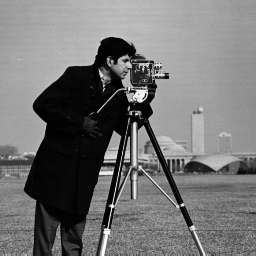} &
\includegraphics[height=\fw]{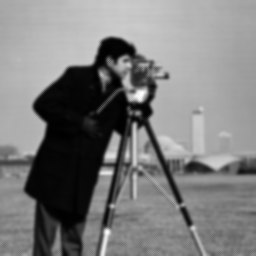} &
\includegraphics[height=\fw]{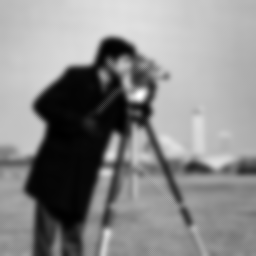} &
\includegraphics[height=\fw]{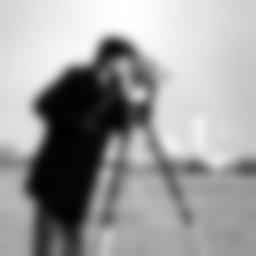} &
\includegraphics[height=\fw]{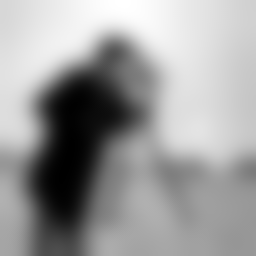} \\
(c) NSS &
\includegraphics[height=\fw]{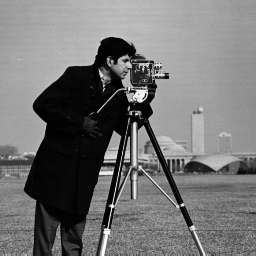} &
\includegraphics[height=\fw]{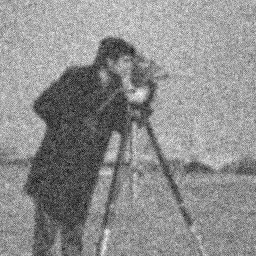} &
\includegraphics[height=\fw]{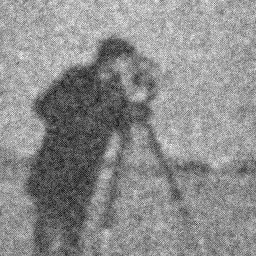} &
\includegraphics[height=\fw]{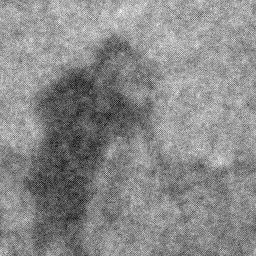} &
\includegraphics[height=\fw]{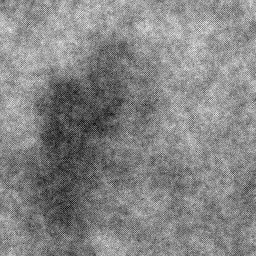} \\
& 
\multicolumn{1}{c}{\quad $t=0$} & 
\multicolumn{1}{c}{\quad $t=4$} & 
\multicolumn{1}{c}{\quad $t=16$} & 
\multicolumn{1}{c}{\quad $t=64$} & 
\multicolumn{1}{c}{\quad $t=256$} & 
\\
\end{tabular}
\caption{A real image in (a) the conventional noise-space (NS) or the repetition of the Gaussian noise with $\sigma=0.15$, 
(b) the conventional scale-space (SS) or the repetition of smoothing, and (c) the proposed noisy scale-space (NSS) (columns) over the time ($t$).}
\label{fig:real_images_with_filters}
\end{figure}
%
%

%
%
\def \rh {0.8} 
\def \fw {100pt}
\begin{figure} [h!]
\centering
\small
\renewcommand{\arraystretch}{\rh}
\begin{tabular}{cccc}
\includegraphics[width=\fw]{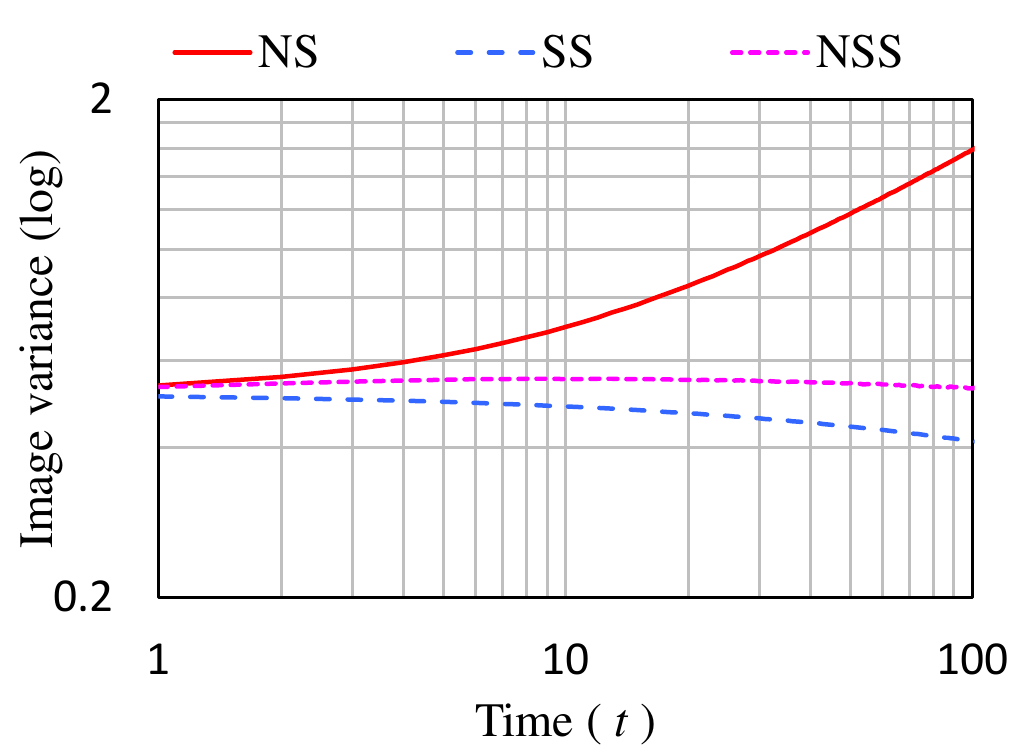} &
\includegraphics[width=\fw]{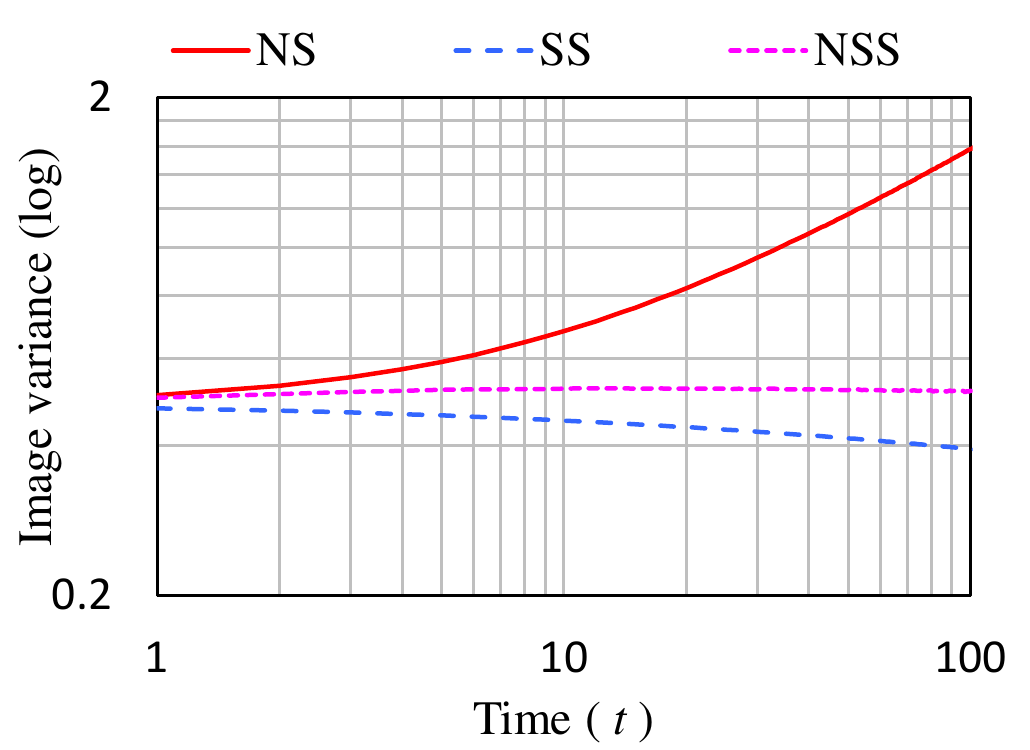} \\
CelebA & Church \\
\includegraphics[width=\fw]{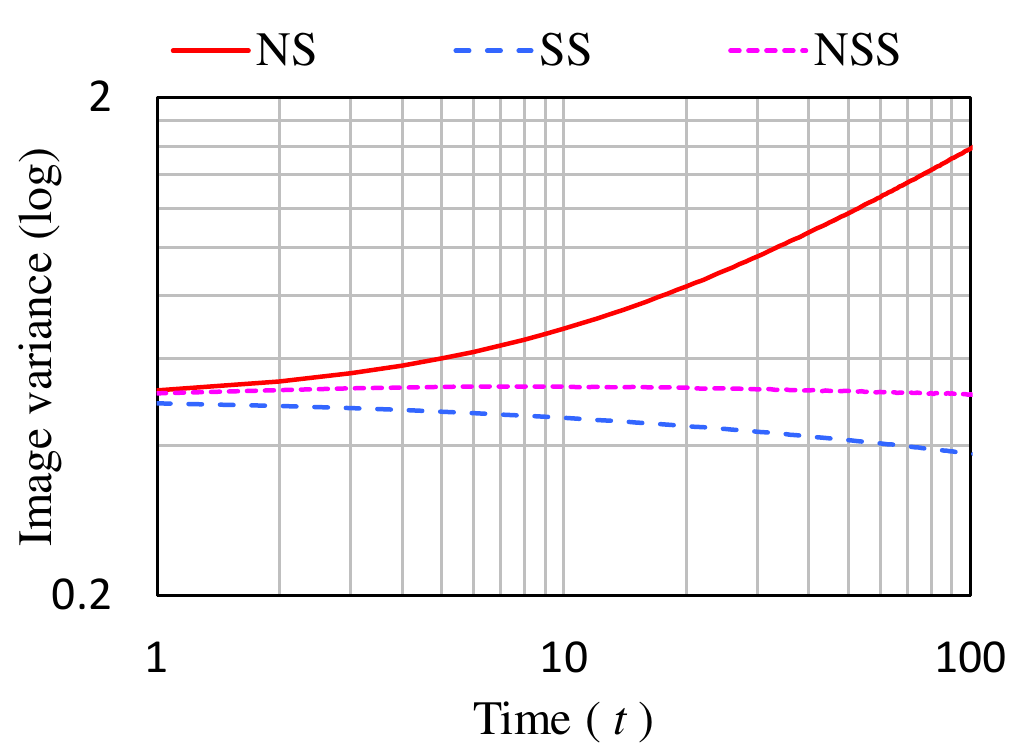} &
\includegraphics[width=\fw]{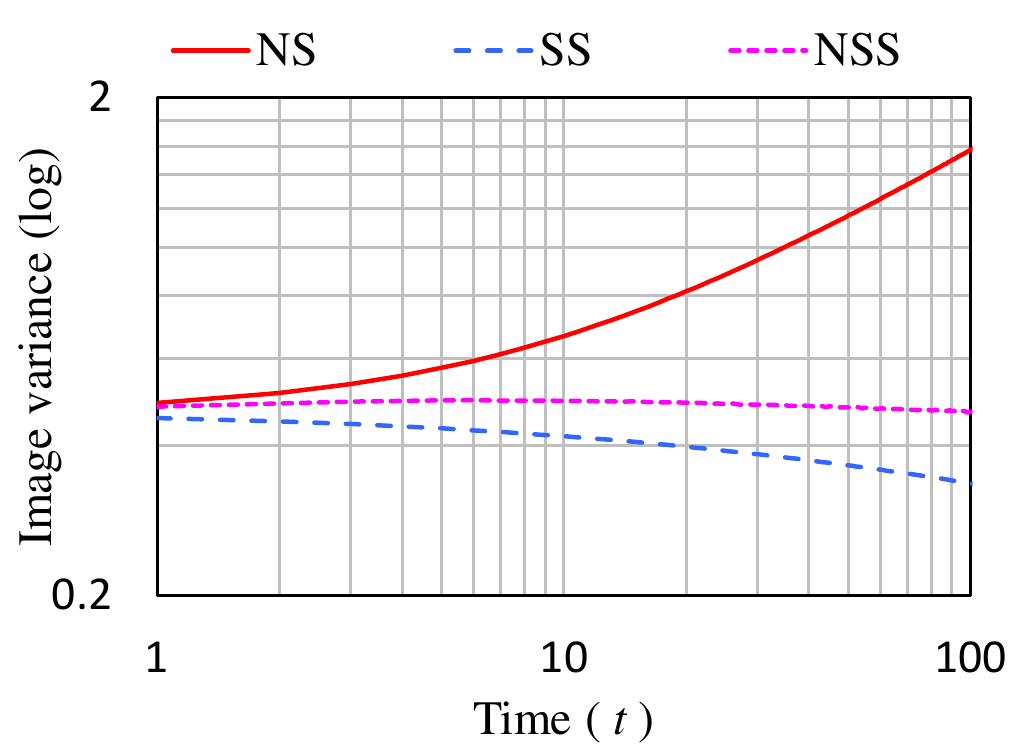} \\
 Conference & Flowers \\
\end{tabular}
\renewcommand{\arraystretch}{1}
\caption{(y-axis) The variance of pixel intensities in the noise-space (red line), the scale-space (blue dashed-line), and the proposed noisy scale-space (magenta dotted-line) over (x-axis) the time $t$ based on 128 images in CelebA~\cite{liu2015faceattributes}, LSUN-Church, LSUN-Conference~\cite{yu15lsun}, and Oxford-Flowers~\cite{Nilsback08flowers} using the noise of $\sigma=0.15$.}
\label{fig:image_variance}
%
%
\end{figure}
%
%
%
%

\section{Stabilization of GAN training via noisy scale-space} \label{sec:method}
\subsection{Proposed noisy scale-space}
We present a data representation, namely noisy scale-space (NSS), that is designed to improve the stability of the optimization for GANs while preserving characteristic features in the generation process.
The presented data representation is a balanced composition of the noise with the diffusion such that we can train GANs using the \textit{smoothed} data that is easy to create by the generator first, while flattening the fake distribution in the high-frequency domain, making the current solution be capable of learning the high-frequency information in the further steps.
%
%
Formally, the noisy scale-space is given as
%
%
\begin{equation} \label{eq:noisy_scale-space}
\Phi_t (y) \coloneqq k \ast \Phi_{t-1} (y) + \epsilon_t,
\end{equation}
%
%
where $k$ is typically the $3 \times 3$ Gaussian kernel, the symbol $\ast$ denotes the convolution, $\epsilon_t \sim N(0,\sigma)$, $\Phi_0 (y)=y$, and $t \in [T]$, i.e., we apply the kernel and the noise to data simultaneously $T$-times.
Figure~\ref{fig:real_images_with_filters} presents an example of image in (top) the conventional noise-space, (middle) the conventional scale-space, and (bottom) the proposed noisy scale-space.
Figure~\ref{fig:image_variance} shows the image variance of the three data-spaces.
\par
The noise variance ($\sigma$) is the primal hyper-parameter that balances the diffusion with the noise in the proposed data representation.
The conventional scale-space (diffusion) removes high-frequency information in image and decreases the image variance as shown in Figure~\ref{fig:image_variance}.
We determine $\sigma$ in the noisy scale-space such that the image variance is almost preserved over $t$.
%
The noisy scale-space is robust to $\sigma$ for natural images as shown in Figure~\ref{fig:image_variance} and we employ $\sigma=0.15$ for $64^2$-pixel images in our experiments.
Moreover $\sigma$ can be determined using only the real images.
%
%
%

%
%
\subsection{Comparison to conventional noise-space}
Given a constant $\sigma$, we can obtain the conventional noise-space as
%
%
\begin{equation} \label{eq:noise-space}
\Phi^\text{n}_t (y) \coloneqq \Phi^\text{n}_{t-1} (y)  + \epsilon_t, 
\end{equation}
%
%
with $\epsilon_{t} \sim  N(0,\sigma), t \in [T]$, and $\Phi^\text{n}_0 (y) = y$.
Equation (\ref{eq:noise-space}) provides a discrete representation of data with additive noises.
Figure~\ref{fig:real_images_with_filters} (a) shows an example of noise-space using a real image. We refer to GAN trained using the noise-space as Noise-Space (NS) GAN.
Considering the sum of normal distributions, the noise-space defined by Eq.(\ref{eq:noise-space}) is equivalent to
\begin{eqnarray} \label{eq:noise-space:closed}
\Phi^\text{n}_t (y) &=& y  + \epsilon_1+ \epsilon_2 + ... + \epsilon_t, \nonumber \\
&=& y  + \hat{\epsilon}(t), 
\end{eqnarray}
where $\hat{\epsilon}(t) \sim N(0, \sigma \cdot t)$, $t \in [T]$. Therefore, the drawback of the noise-space is that it increases the variance of data, and makes the original data harder to be learned.
\par
Note that the proposed noisy scale-space using Eq.(\ref{eq:noisy_scale-space}) can be rewritten in a closed-form as
%
%
\begin{eqnarray} \label{eq:noisy_scale-space:closed}
\Phi_t (y) &=& 
\underbrace{k \ast^{(t)} y}_{\textrm{smoothed data}} +  
\underbrace{
k \ast^{(t-1)} \epsilon_{1} + k \ast^{(t-2)} \epsilon_{2} + ... + k \ast^{(1)} \epsilon_{t-1} + k \ast^{(0)} \epsilon_{t} 
}_{\textrm{low-to high-frequency noises}}, \nonumber \\
\end{eqnarray}
where $k \ast^{(t)}$ denotes the $t$-times convolution with kernel $k$, and $k\ast^{(0)}\epsilon_{t}=\epsilon_{t}$.
Thus, our noisy scale-space balances the variances of the data-term and the noise-terms using the smoothing. A finding is that we can keep the data variance using a constant $\sigma$ as demonstrated in Figure~\ref{fig:image_variance}.
Equation (\ref{eq:noisy_scale-space:closed}) also shows that the noisy scale-space replaces the high-frequency information in data by a set of noises with the corresponding frequencies. 
We demonstrate this property using a synthetic dataset in the following section.
%
%
%
%
%
%
%

%
%
%
%
\def \fw {58pt}
\def \pw {58pt}
\begin{figure}[htb]
\centering
\scriptsize
\begin{tabular}{ccccc}
\includegraphics[width=\fw]{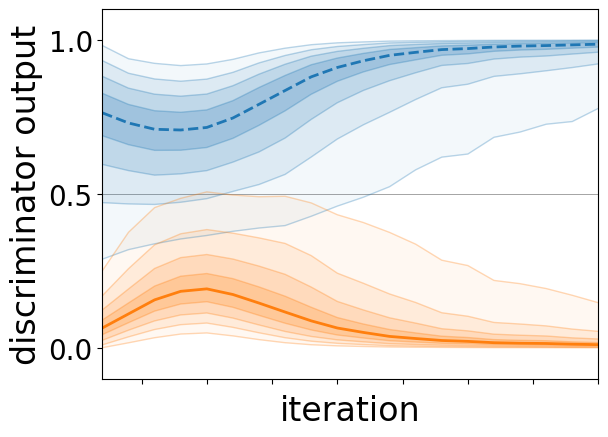} &
\includegraphics[width=\fw]{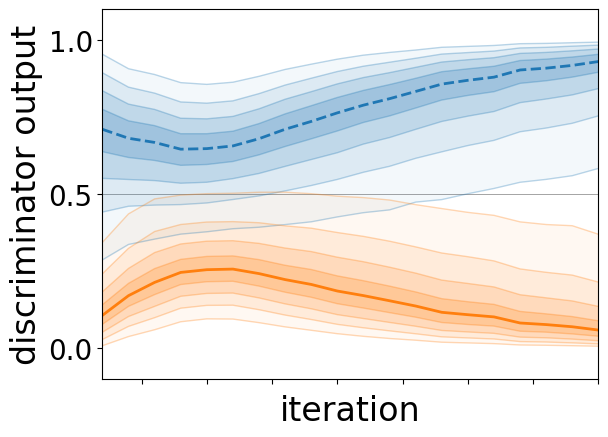} &
\includegraphics[width=\fw]{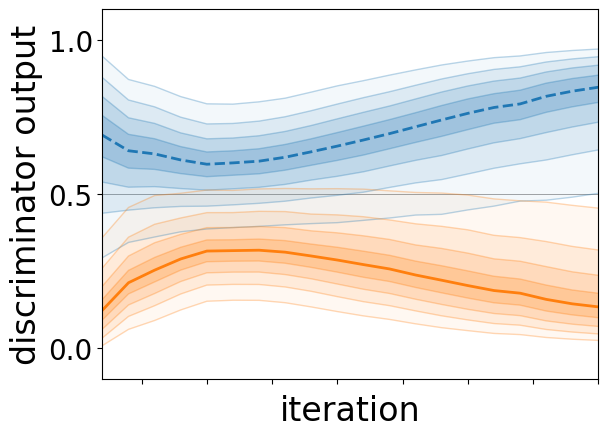} &
\includegraphics[width=\fw]{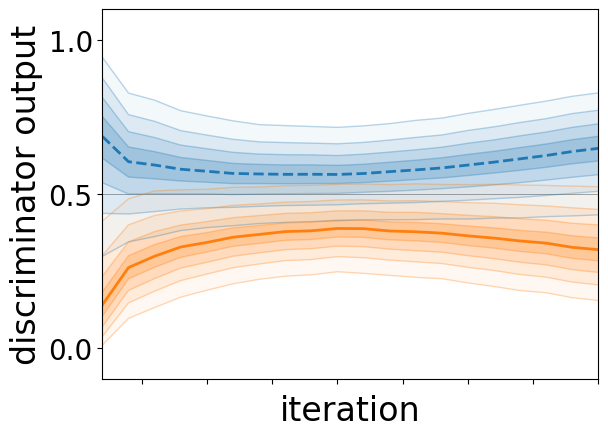} &
\includegraphics[width=\fw]{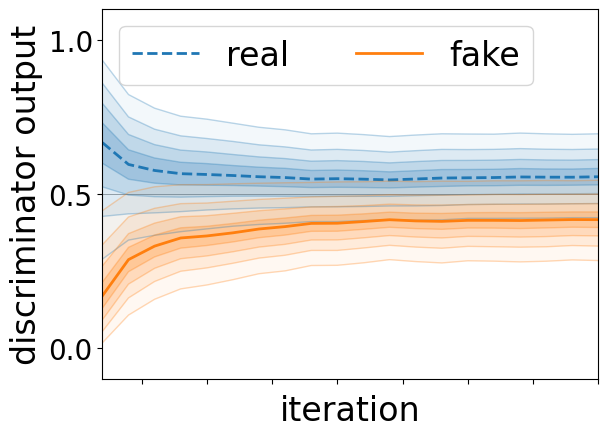}  \\
\multicolumn{1}{c}{$t=0$} &
\multicolumn{1}{c}{$t=8$} &
\multicolumn{1}{c}{$t=16$} &
\multicolumn{1}{c}{$t=32$} &
\multicolumn{1}{c}{$t=64$}  \\
\end{tabular}
\caption{
The prediction curves by discriminator for (blue doted line) the projected real data and (orange line) the projected fake data 
by DCGAN trained using CelebA dataset with (1st, ..., 5th columns) the fixed smoothing with $t$: $t=0$ is the baseline. 
The percentiles of discriminator output were visualized in epoch wise.}
\label{fig:preliminary:smoothedGAN}
\end{figure}
%
%

%
%
%
%
\begin{table}[htb]
\centering
\scriptsize
\caption{
Fr\'echet inception distance (FID) and inception score (IS) for Celeba dataset by DCGAN trained using (1st, ..., 7th columns) two-step scale-space $\{t,0\}$ with equal periods using $t=0,2,...,128$, respectively, and (8th column) a scale-space using exponential annealing with $t=128$: $t=0$ is the baseline. The mean of FID and IS were computed within 20 trials.}
\label{tab:FID:smoothing}
\begin{tabular}{l | ccccccccc}
\hline
Smoothing          & $t=0$ & $t=4$ & $t=8$ & $t=16$ & $t=32$ & $t=64$ & $t=128$ & Annealing \\
\hline
FID ($\downarrow$) & \bf{21.36}  & 21.94 & 24.70 & 25.06  & 25.29  & 26.87  & 60.34   & 21.49  \\
IS ($\uparrow$)    & 2.39   & 2.39  & 2.35  & 2.33   & 2.30   & 2.31   & 2.14    & 2.37  \\
\hline
\end{tabular}
\end{table}
%
%

%
\subsection{Comparison to conventional scale-space} 
The conventional scale-space is a multi-scale representation of image in which the smoothing is applied to data recursively as
%
%
\begin{equation} \label{eq:scale-space}
\Phi_t^\text{s} (y) \coloneqq k \ast \Phi^\text{s}_{t-1}(y) , 
\end{equation}
%
%
with $\Phi^\text{s}_0 (y)=y$ and $t \in [T]$.
As shown in Figure~\ref{fig:real_images_with_filters} (b), the scale-space continuously removes the low-level information, e.g., textures and details,
and provides the high-level information that is invariant to the scale. 
%
%
%
\par

We first re-examine the effect of diffusion in GAN optimization. Figure~\ref{fig:preliminary:smoothedGAN} visualizes the prediction curves of DCGAN~\cite{radford2015dcgan} based on CelebA~\cite{liu2015faceattributes} where we trained the networks using the fixed smoothing time ($t$) applied to both real and fake data. 
We have chosen DCGAN since the convolution networks is the essential architecture of the modern GANs, e.g.,~\cite{chen2016infogan,pathakCVPR16context,odena2017conditional,zhang2017stackgan,CycleGAN2017,brock2018large,karras2019style}.
As demonstrated in Figure~\ref{fig:preliminary:smoothedGAN}, a larger $t$ results in better prediction curves where both $D(\Phi^\text{s}_t(x))$ and $D(\Phi^\text{s}_t(G(z)))$ converge to 0.5. Thus, the diffusion will make the problem more easy to learn by GANs. 
Note that the quality of $G(z)$ degenerates with $t$ since there is information loss due to the diffusion. Therefore, it is natural to anneal $t$ over the training process. 
\par
However, the conventional scale-space cannot improve the accuracy of GANs in general.
Table~\ref{tab:FID:smoothing} summaries the Fr\'echet inception distance (FID)~\cite{Heusel17TTUR} and the inception score  (IS)~\cite{Sailmans16improved} by DCGAN for CelebA~\cite{liu2015faceattributes} using scale-spaces, where we first used a fixed smoothing $t$ for 5 epochs then used the original data $(t=0)$ for 5 epochs. More details about the experimental set-up are summarized in Section~\ref{sec:subsec:experimental-setup}. 
As shown in Table~\ref{tab:FID:smoothing}, GANs using scale-spaces are often inferior to the baseline without smoothing.
%
\par
Our key observation is that the scale-space has a side effect in the GAN optimization 
where the smoothed data (real and fake data) make the generator create a smoothed data and thus shrink the capability of the generator to learn high-frequency information in further training steps.
To demonstrate this side effect, we propose to visualize the true probability distributions of both reals and fakes using a synthetic dataset that is created based on eight of vertical Hadamard bases ($\{B_i\}^8$) shown in Figure~\ref{fig:Hadamard_coefficients} (top).  
Given coefficients $\{\alpha_i\}^8$ for the eight basis,  $\sum_i \alpha_i B_i$ produces an $8\times8$ pixel image as real data that looks a white/grey vertical stripes.
The coefficients are the true probability distribution of real data.
Given fake data, we can fit the bases to the fake image and compute the coefficients of eight bases that reflect the probability distribution of fakes with the fitting residuals.
%
%
%
%
%
%

%
%
%
%
\def \fw {70pt}
\def \gw {150pt}
\begin{figure}[h!]
\centering
\small
%
%
%
\includegraphics[width=300pt]{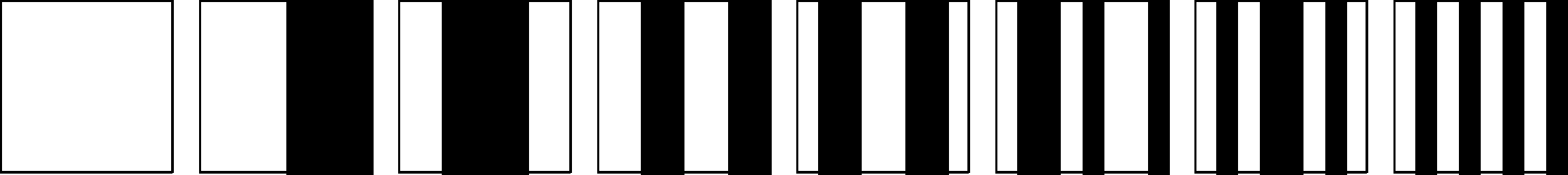} \\
\vspace{8pt}
\begin{tabular}{cccc}
\includegraphics[width=\fw]{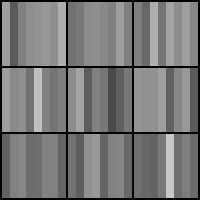} &
\includegraphics[width=\fw]{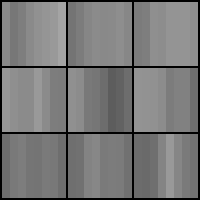} & 
\includegraphics[width=\fw]{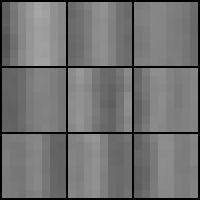} &
\includegraphics[width=\fw]{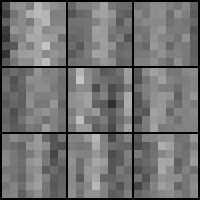} \\
$\real$ & 
$\Phi^\text{s}(\real)$ &
$\Phi^\text{s}(\fake)$ & 
$\fake$ \\
\end{tabular} \\
\begin{tabular}{cc}
\includegraphics[width=\gw]{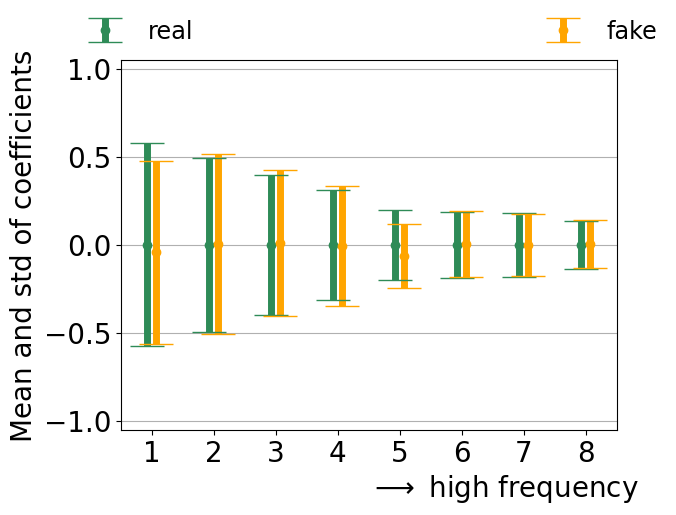} &
\includegraphics[width=\gw]{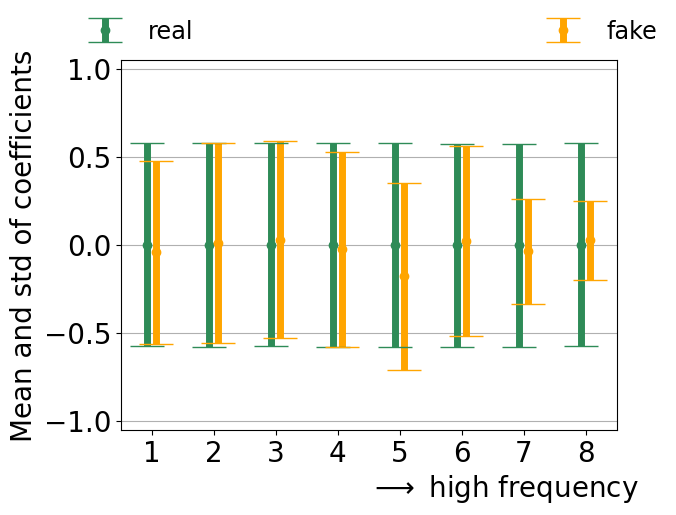}  \\
$\{\alpha_i\}$ of $\Phi^\text{s}(\real), \Phi^\text{s}(\fake)$ &  
$\{\alpha_i\}$ of $\real, \fake$ \\
\end{tabular} \\
\caption{GAN with synthetic data using the scale-space:
(top) The Hadamard bases ($B_1, ..., B_8$), 
(middle part) 
nine examples of the original reals ($\real$), 
the smoothed reals ($\Phi^\text{s}(\real)$) with fixed $t=8$, 
the smoothed fakes ($\Phi^\text{s}(\fake)$) with fixed $t=8$, 
and the original fakes ($\fake$), 
(bottom left) (y-axis) the mean and std. of the coefficients for (x-axis) the Hadamard bases $B_1,...,B_8$ with low to high frequencies within the smoothed reals (green) and the smoothed fakes (orange), 
and (bottom right) those within the original reals and fakes.
$200$K of the synthetic images were created using $\alpha_i \sim U(-1,1), \forall i$ with the uniform distribution ($U$). 
We applied the smoothing to all data in the batch.
The basic GAN~\cite{goodfellow2014generative} was trained using Adam with the learning-rate scale of $\eta=2 \times 10^{-5}$ for 200 epochs.
}
\label{fig:Hadamard_coefficients}
\vspace{4pt} 
\end{figure}
%
%
%
%
%
\def \gw {150pt}
\begin{figure}[h!]
\centering
\small
%
%
\begin{tabular}{cc}
\includegraphics[width=\gw]{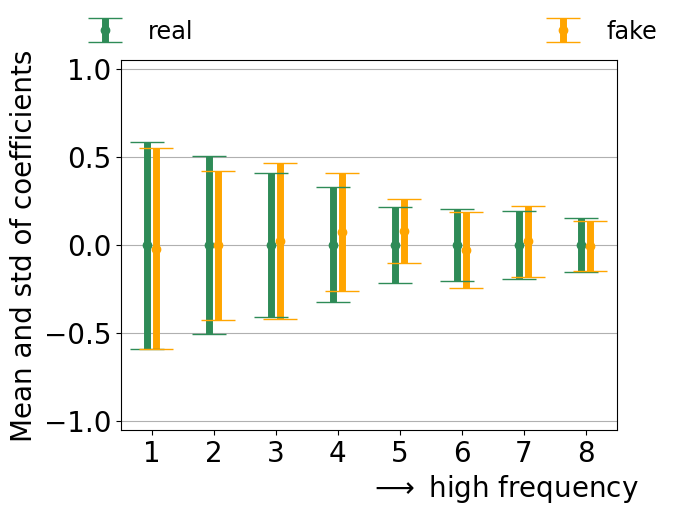} &
\includegraphics[width=\gw]{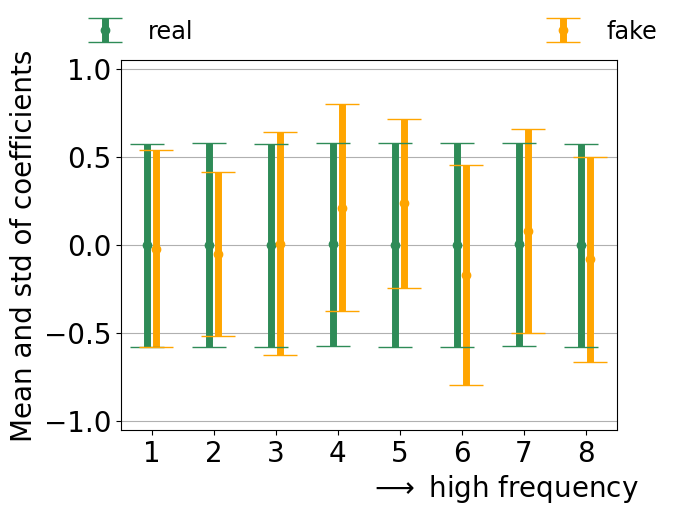} \\
$\{\alpha_i\}$ of $\Phi(\real), \Phi(\fake)$ & $\{\alpha_i\}$ of $\real, \fake$ \\
\end{tabular} \\
\caption{GAN with synthetic data using the proposed noisy scale-space: (left) The mean and std. of  the Hadamard coefficients (y-axis) with low to high frequencies (x-axis) within the projected reals (green) and the projected fakes (orange), and (right) those within the original reals and fakes using the noisy scale-space with fixed $t=8$.}
\label{fig:Hadamard_coefficients_smoothing-with-noise}
\end{figure}
%
%

%
%
%
\par
We trained the basic GAN~\cite{goodfellow2014generative} based on the synthetic data using the smoothing with fixed $t=8$.
Figure~\ref{fig:Hadamard_coefficients} illustrates 
(top) the Hadamard bases,
(middle) example of the original reals, 
smoothed-reals, smoothed-fakes, and the original fake data, and 
(bottom) the distribution of their Hadamard coefficients.
There are two observations:
(bottom-left) The coefficients of smoothed fakes $\Phi(\fake)$ follow the those of $\Phi(x)$ of which high-frequency coefficients are suppressed by the diffusion, i.e., the model learned the probability distribution of the smoothed data.
(bottom-right) However, the diffusion decreased the diversity of high-frequency coefficients of fakes ($\fake$).
This means that the data smoothing reduces the overlap between the fake distribution with the distribution of reals with fine details that will be given in further steps.
\par
We visualize the effect of the noisy scale-space using the synthetic dataset in Figure~\ref{fig:Hadamard_coefficients_smoothing-with-noise} 
in which we used a fixed $t=8$.
Figure~\ref{fig:Hadamard_coefficients_smoothing-with-noise} shows that 
(left) the projected data are smoothed yet (right) the diversity of high-frequency coefficients of fakes are preserved as expected.
\par
Here, we have studied the side-effect of the diffusion.
Interestingly, this kind of phenomenon due to a data transformation in GAN optimization is called 'leaking' that has been studied on, e.g., rotation and hue change~\cite{karras2020training}. 
The contributions of our work are that we visualize the leaking of the smoothing using the synthetic data; and we propose the prescription to mitigate the side-effect of diffusion in GAN training.
\def \fwA{160pt}
\def \fwB{75pt}
\begin{figure} [htb!]
%
%
\centering
\small
\begin{tabular}{m{\fwA} m{\fwB} m{\fwB}}
\includegraphics[width=\fwA]{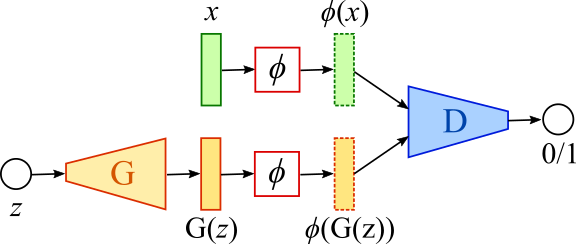}&
\includegraphics[width=\fwB]{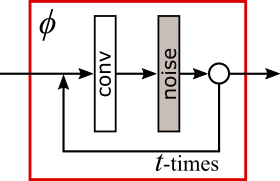}&
\includegraphics[width=\fwB]{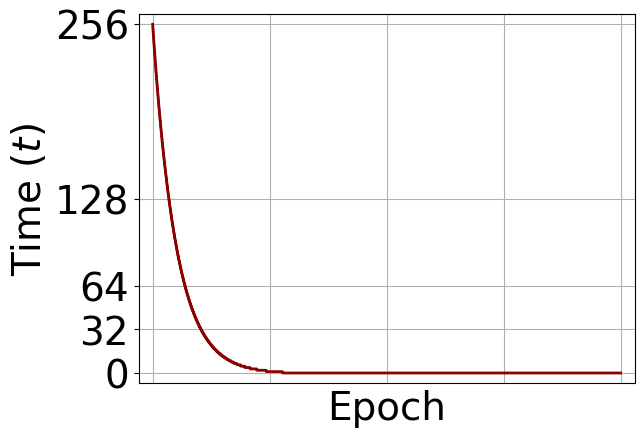} \\
\multicolumn{1}{c}{(a)} &
\multicolumn{1}{c}{(b)} &
\multicolumn{1}{c}{\qquad (c)} \\
\end{tabular}
\caption{(a) The pipeline of GAN training with function $\Phi$ that is applied to the half of both real data ($\real$) and fake data ($\fake$); (b) The module of the noise scale-space that consists of $t$-times repetition of the convolution-layer using the Gaussian kernel with the random noise-layer; and (c) The exponential annealing of $t$ using the power of $\beta=20$ with $T = 256$.}
\label{fig:architecture_of_SSN-GAN}
%
%
\end{figure}
%
%

%
\subsection{Implementation of noisy scale-space}
The data-based stabilization methods can be embedded into GANs as depicted in Figure~\ref{fig:architecture_of_SSN-GAN} (a).
We implement our data representation into DCGAN~\cite{radford2015dcgan} that is the foundation of recent extensive studies, and call it Noisy Scale-Space (NSS) GAN.
The function $\Phi$ consists of the repetition of the smoothing and the noise-injection layers (Figure~\ref{fig:architecture_of_SSN-GAN}b) that can be computed efficiently in parallel.
Regarding the annealing of the time parameter ($t$) that determines the magnitude of stabilization effect, we use an exponential function as
%
%
\begin{equation} \label{eq:time-annealing}
t^{(i)} \coloneqq T \cdot \exp \left( i/\beta \right),
\end{equation}
%
%
where $i \in [0,1]$ is the relative iteration in the optimization process that starts at $i=0$, $\beta$ is the power of decay, and $T$ is the initial time $t^{(0)}$. 
We use $\beta=20$ in order to apply the data transformation in the early stage of GAN optimization.
Figure~\ref{fig:architecture_of_SSN-GAN} (c) shows the annealing curve of $t$ with $T=256$.
We have observed that the exponential function achieves better accuracy than others including the step function and the decaying-wave function.
Also we employ an implementation technique~\cite{jenni2019noiseGAN} in which we apply the filtering function $\Phi$ to only half of data in real-and fake-batches for obtaining a stable and accurate generator.
%
%
%

\section{Experimental results} \label{sec:results}
In order to empirically demonstrate the expected stabilization effect of the noisy scale-space for GAN optimization, we conduct three experiments:
an ablation study on the initial scale in data spaces,
a primal experiment in which we compare the proposed NSS with the state-of-the-art GANs based on DCGAN,
and an additional experiment using StyleGAN2.
\subsection{Experimental set-up for primal experiments} \label{sec:subsec:experimental-setup}
In the preliminary experiment, we compare our NSS-GAN with potential competitors:
DCGAN (GAN) as the baseline,  GAN using the scale-space (SS-GAN), DCGAN using the noise-space (NS-GAN) where SS-GAN, NS-GAN, and NSS-GAN share the same architecture and the annealing function of $t$ except for the filtering function $\Phi$.
Moreover, we employ GAN with the gradient regularization~\cite{Mescheder2018which} (GAN-gr), LSGAN~\cite{mao2017LSGAN}, WGAN-gp~\cite{Sailmans16improved}, and Dragan~\cite{kodali2017convergence} as the state-of-the-arts in the second experiment.
\par%
We use four of the major datasets: 
CelebA~\cite{liu2015faceattributes}, 
LSUN-Church, LSUN-Conference~\cite{yu15lsun}, 
and Oxford-Flowers~\cite{Nilsback08flowers}
as the image generation tasks of faces, outdoor scenes, indoor scenes, and plants, respectively.
CelebA consists of about 200K of celebrity face images. LSUN-Church and LSUN-Conference have about 126K images of outdoor scene of churches, 
and about 224K images of indoor scene of conferences, respectively. Oxford-Flowers has about 8K images of flowers. 
%
The images are resized to $64\times64$ pixels to meet the architecture of DCGAN.
\par
We use the non-saturating loss that is known to be superior to the Wasserstein loss and others~\cite{Mescheder2018which}, and employ Adam~\cite{kingma2014adam} as one of the popular optimizers in GAN studies.
For all the experiments, the mini-batch size is set to 128, and the number of training epochs is set to 10 epochs for CelebA, LSUN-Church and LSUN-Conference, and 100 epochs for Oxford-Flowers, respectively, based on their example sizes.
The standard deviation of noise is set to $\sigma=0.15$.
The hyper-parameters that determine the quality of generated data are 
the initial time $T$ of SS, NS, and NSS-GANs,
the learning-rate scale ($\eta$) and the first momentum coefficient ($b_1$) of Adam,
and the regularization coefficient ($\lambda$) of GAN-gr.
We conduct our experiments in two steps: a preliminary experiment on $T$ with fixed $\eta$ and $b_1$, and the final experiment using the tuned $\eta$, $b_1$, and $\lambda$ with fixed $T$.
We perform each condition 20 individual times.
\par
For quantitative evaluation of generated fake images, we use the Fr\'echet inception distance (FID)~\cite{Heusel17TTUR} 
with the inception score (IS)~\cite{Sailmans16improved} that are widely used in GAN studies.
FID measures the distance between the real data with fake data in feature space defined using a pre-trained network.
IS measures the diversity of fakes in the feature space.
Lower FID values with higher IS indicate better quality and diversity of fake data, respectively.
We use FID as the primal metric that reflects the objective of GAN.
%
%
%

%
%
%
%
\def \rh {1.2} 
\def \pw {12pt}
\def \qw {16pt}
\begin{table}[htb]
%
%
\caption{The Fr\'echet inception distance (FID) over the initial time $T=0,32,64,128,256$ for (column parts) CelebA, LSUN-Church, LSUN-Conference, and Oxford-Flowers by DCGAN using the scale-space (SS), the noise-space (NS), and our noisy scale-space (NSS): In order to demonstrate the stabilization effect over $T$, the learning-rate scale and the 1st momentum were fixed for each dataset such that the baseline DCGAN ($T=0$) can be unstable.
}
\label{tab:tuning-T}
\centering
\scriptsize
\renewcommand{\arraystretch}{\rh}
\begin{tabular}{p{20pt} | P{\pw}P{\pw}P{\qw} | P{\pw}P{\pw}P{\qw} | P{\qw}P{\pw}P{\qw} | P{\qw}P{\pw}P{\qw}}
\hline
& \multicolumn{3}{c|}{CelebA}  & \multicolumn{3}{c|}{Church}   & \multicolumn{3}{c|}{Conference} &  \multicolumn{3}{c}{Flowers} \\ 
& SS	& NS	& NSS	& SS	& NS	& NSS	 & SS	& NS	& NSS	 & SS	& NS	& NSS	 \\ \hline
$T$=0     & 24.60	& -	 & -	& 70.94	& -	 & -	& 127.32	& -	& -	& 201.02	& -	  & -	\\ \hline
$T$=32    & 23.14	& 22.82& 22.32	& 69.80	& 70.20	& 69.88	& 119.06	& 71.16	& 67.37 & 162.24	& 89.46	& 87.30 \\ \hline
$T$=64    & 23.16	& 23.01	& 22.43	& 69.05	& 69.60	& 69.67	& 98.73	 & 72.21	& 70.33	& 120.57	& 93.46	& 88.68	\\ \hline
$T$=128   & 23.11	& 24.02	& 21.18 & 69.38	& 69.51	& 68.59	& 84.31	 & 71.83	& 68.90	& 116.29	& 95.66	& 92.57	\\ \hline
$T$=256   & 22.31	& 24.29	& 21.79	& 70.58	& 69.61	& 67.38 & 87.46	 & 73.11	& 70.47	& 100.84	& 95.47	& 94.24	\\ 
\hline
\end{tabular}
\renewcommand{\arraystretch}{1}
%
\end{table}
%
%
%
%
%
%
%
\def \rh {0.8} 
\def \pw {22pt}
\def \qw {38pt}
\begin{table}[htb]
\caption{The tuned hyper-parameters of (columns) the experimented GANs for (rows) CelebA, LSUN-Church, LSUN-Conference, and Oxford-Flowers datasets: the learning-rate scale ($\eta \times 10^{-4}$ ), the 1st momentum ($b_1$) of Adam, and the regularization coefficient ($\lambda$) were selected using grid search based on the mean FID. GAN with the scale-space (SS), GAN with the noise-space (NS), and GAN with the noisy scale-space (NSS) share the fixed $T=256$.}
\label{tab:tuned_condition}
\scriptsize
\centering
\renewcommand{\arraystretch}{\rh}
\begin{tabular}{l l P{\pw}P{\qw}P{\pw}P{\qw}P{\pw}P{\pw}P{\pw}P{\pw}}          
\hline
 &  & GAN  & GAN-gr  & LSGAN  & WGAN-gp  & Dragan  & SS  & NS  & NSS \\
\hline
\parbox[t]{2mm}{\multirow{3}{*}{\rotatebox[origin=c]{90}{CelebA}}}  & $\eta$  & 5  & 5  & 1  & 10  & 1  & 5  & 5  & 5 \\
  & $b_1$  & 0.3  & 0.3  & 0.4  & 0.3  & 0.4  & 0.4  & 0.3  & 0.4 \\
  & $\lambda$  & -  & 1  & -  & 20  & 10  & -  & -  & - \\
\hline
\parbox[t]{2mm}{\multirow{3}{*}{\rotatebox[origin=c]{90}{Church}}}  & $\eta$  & 2  & 5  & 0.5  & 10  & 1  & 5  & 10  & 5 \\
  & $b_1$  & 0.3  & 0.3  & 0.5  & 0.3  & 0.4  & 0.3  & 0.3  & 0.3 \\
  & $\lambda$  & -  & 10  & -  & 20  & 10  & -  & -  & - \\
\hline
\parbox[t]{2mm}{\multirow{3}{*}{\rotatebox[origin=c]{90}{Confer.}}}  & $\eta$  & 2  & 2  & 0.5  & 10  & 2  & 2  & 5  & 5 \\
  & $b_1$  & 0.3  & 0.3  & 0.5  & 0.3  & 0.4  & 0.3  & 0.3  & 0.3 \\
  & $\lambda$  & -  & 5  & -  & 20  & 10  & -  & -  & - \\
\hline
\parbox[t]{2mm}{\multirow{3}{*}{\rotatebox[origin=c]{90}{Flowers}}}  & $\eta$  & 2  & 5  & 0.5  & 10  & 2  & 5  & 5  & 5 \\
  & $b_1$  & 0.4  & 0.4  & 0.5  & 0.3  & 0.5  & 0.3  & 0.3  & 0.3 \\
  & $\lambda$  & -  & 5  & -  & 20  & 10  & -  & -  & - \\
\hline
\end{tabular}          
\renewcommand{\arraystretch}{1}
%
%
\end{table}
%
%
%
%

%
%
%
%
\subsection{Effect of initial time}
We first examine the initial time $T$ in Eq.(\ref{eq:time-annealing}) that determines the degree of the scale-space, the noise-space, and the proposed noisy scale-space.
In order to observe the relative stabilization effect of the three data representations in comparison to the baseline GAN, we purposely use a condition that can make the baseline GAN unstable for each dataset. Concretely, we employ $(\eta,b_1)=(0.0002,0.4)$ for CelebA and LSUN-Church, $(\eta,b_1)=(0.0005,0.3)$ for LSUN-Conference, and $(\eta,b_1)=(0.0005,0.4)$  for Oxford-Flowers datasets.
\par
Table~\ref{tab:tuning-T} shows the mean FID within the 20 trials over the initial time of $T = 0,32,64,128,256$ where $T = 0$ is the baseline DCGAN.
Table~\ref{tab:tuning-T} demonstrates that both the scale-space and the noise-space improved the quality of fakes compared to the baseline GAN; and the proposed NSS-GAN has achieved a better stabilization effect than SS-GAN and NS-GAN independent of the datasets and the initial time $T$.
%
The scale ($t$) should start with a large value for any type of images and the initial value $T$ depends on the context and the scale of images essentially. In practice, we recommend $T = 256$ for $64^2$-pixel images and $T = 128$ for $512$-pixel images and use them in the final experiments.
%
%
%
%

%
\def \fw {80pt}
\def \pw {75pt}
\begin{figure} [htb!]
\centering
\hspace*{-20pt} 
\def\arraystretch{0.70} 
\vspace{24pt} 
\begin{tabular}{P{\pw}P{\pw}P{\pw}P{\pw}}
\includegraphics[width=\fw]{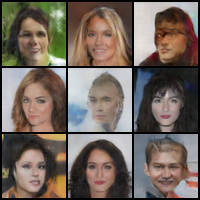} &
\includegraphics[width=\fw]{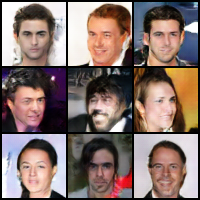} &
\includegraphics[width=\fw]{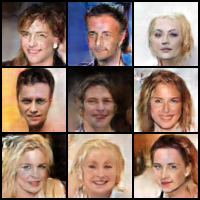} &
\includegraphics[width=\fw]{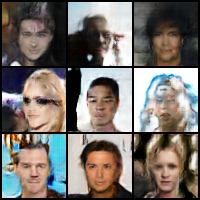} \\
\quad GAN & \quad GAN-gr & \quad LSGAN & \quad WGAN-gp\\ 
\includegraphics[width=\fw]{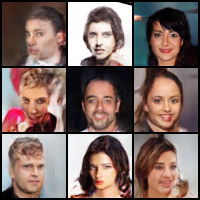} &
\includegraphics[width=\fw]{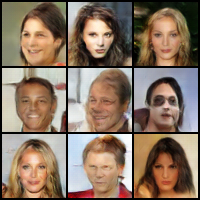} &
\includegraphics[width=\fw]{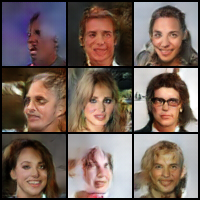} &
\includegraphics[width=\fw]{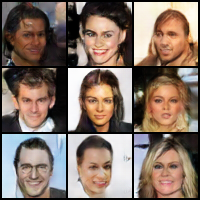} \\
\quad Dragan & \quad SS-GAN & \quad NS-GAN & \quad NSS-GAN \\ 
\multicolumn{4}{c}{(a) CelebA} \vspace{6pt} \\

\includegraphics[width=\fw]{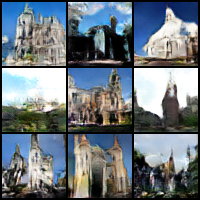} &
\includegraphics[width=\fw]{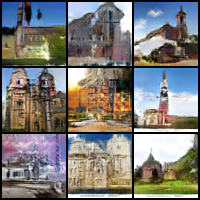} &
\includegraphics[width=\fw]{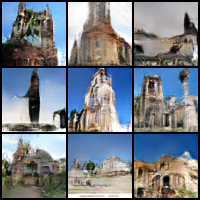} &
\includegraphics[width=\fw]{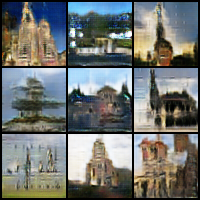} \\
\quad GAN & \quad GAN-gr & \quad LSGAN & \quad WGAN-gp\\ 
\includegraphics[width=\fw]{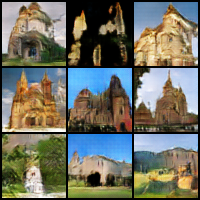} &
\includegraphics[width=\fw]{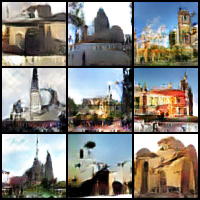} &
\includegraphics[width=\fw]{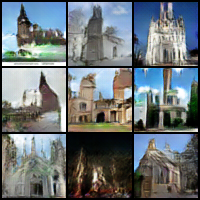} &
\includegraphics[width=\fw]{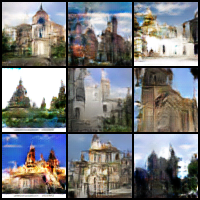} \\
\quad Dragan & \quad SS-GAN & \quad NS-GAN & \quad NSS-GAN \\ 
\multicolumn{4}{c}{(b) Church}
\end{tabular}
\caption{Fake images created by the generator of which FID is the closest to the mean of the individual trials for 
(top part) CelebA and (bottom part) LSUN-Church datasets using
the baseline DCGAN (GAN), 
GAN with gradient regularization (GAN-gr)~\cite{Mescheder2018which}, 
LSGAN~\cite{mao2017LSGAN},
WGAN-gp~\cite{Sailmans16improved},
(bottom)
Dragan~\cite{kodali2017convergence}, 
and GANs using the scale-space (SS), 
the noise-space (NS), 
and the proposed noisy scale-space (NSS) with $T=256$.}
\label{fig:final-comparison-images-part1}
%
\end{figure}
%
%
%
\def \fw {80pt}
\def \pw {75pt}
\begin{figure} [h!]
\centering
\hspace*{-20pt} 
\def\arraystretch{0.70} 
\vspace{24pt} 
\begin{tabular}{P{\pw}P{\pw}P{\pw}P{\pw}}
\includegraphics[width=\fw]{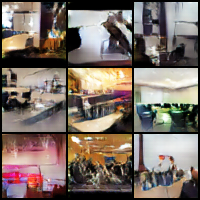} &
\includegraphics[width=\fw]{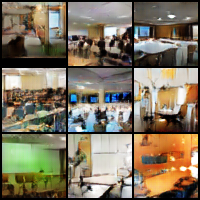} &
\includegraphics[width=\fw]{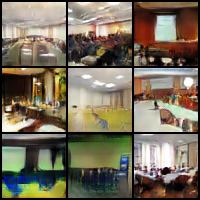} &
\includegraphics[width=\fw]{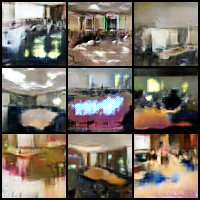} \\
\quad GAN & \quad GAN-gr & \quad LSGAN & \quad WGAN-gp\\ 
\includegraphics[width=\fw]{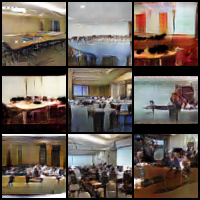} &
\includegraphics[width=\fw]{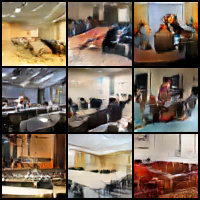} &
\includegraphics[width=\fw]{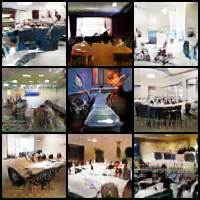} &
\includegraphics[width=\fw]{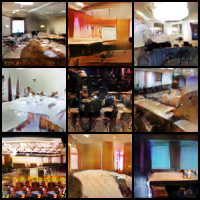} \\
\quad Dragan & \quad SS-GAN & \quad NS-GAN & \quad NSS-GAN \\ 
\multicolumn{4}{c}{(a) Conference} \vspace{6pt} \\

\includegraphics[width=\fw]{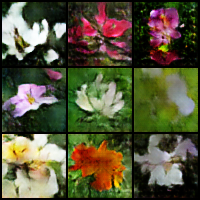} &
\includegraphics[width=\fw]{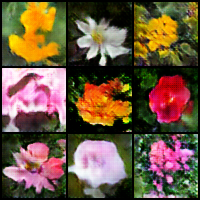} &
\includegraphics[width=\fw]{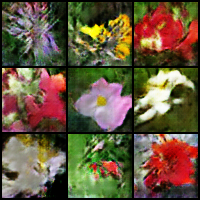} &
\includegraphics[width=\fw]{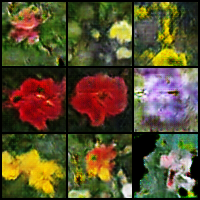} \\
\quad GAN & \quad GAN-gr & \quad LSGAN & \quad WGAN-gp\\ 
\includegraphics[width=\fw]{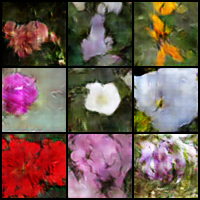} &
\includegraphics[width=\fw]{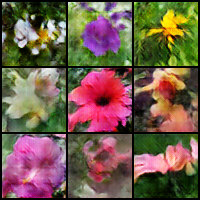} &
\includegraphics[width=\fw]{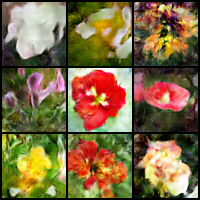} &
\includegraphics[width=\fw]{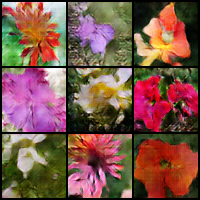} \\
\quad Dragan & \quad SS-GAN & \quad NS-GAN & \quad NSS-GAN \\ 
\multicolumn{4}{c}{(b) Flowers}
\end{tabular}
\caption{Fake images created by the generator of which FID is the closest to the mean of the individual trials for 
(top part) LSUN-Conference and (bottom part) Oxford-Flowers datasets using
the baseline DCGAN (GAN), 
GAN with gradient regularization (GAN-gr)~\cite{Mescheder2018which}, 
LSGAN~\cite{mao2017LSGAN},
WGAN-gp~\cite{Sailmans16improved},
(bottom)
Dragan~\cite{kodali2017convergence}, 
and GANs using the scale-space (SS), 
the noise-space (NS), 
and the proposed noisy scale-space (NSS) with $T=256$.}
\label{fig:final-comparison-images-part2}
%
\end{figure}
%
%
%
\def \rh {1.2} 
\def \pw {61pt}
\begin{table}[h!]         
\caption{The mean and std. of Fr\'echet inception distance (FID) and inception score (IS) within 20 individual trials using CelebA, LSUN-Church, LSUN-Conference, and Oxford-Flowers datasets by (rows in each part) 
the baseline DCGAN (GAN),
GAN with gradient regularization (GAN-gr)~\cite{Mescheder2018which},
LSGAN~\cite{mao2017LSGAN},
WGAN-gp~\cite{Sailmans16improved},
Dragan~\cite{kodali2017convergence}, 
and GANs using the conventional scale-space (SS), the noise-space (NS), and the proposed noisy scale-space (NSS) with
$T=256$: The learning-rate and 1st momentum were tuned for each condition based on the mean FID.}         
\label{tab:final-comparison-metrics}
\centering         
\small
\renewcommand{\arraystretch}{\rh}
\begin{tabular}{l |P{\pw}P{\pw}  |P{\pw}P{\pw}}      
\hline      
 & \multicolumn{2}{c|}{CelebA}  & \multicolumn{2}{c}{Church}   \\
 & FID ($\downarrow$) & IS ($\uparrow$)  & FID & IS  \\
 \hline
GAN	 & 21.36 $\pm$ 1.71	 & 2.39 $\pm$ 0.06	 & 69.05 $\pm$ 4.65	 & 2.98 $\pm$ 0.07		\\
GAN-gr	 & 19.87 $\pm$ 1.19	 & 2.35 $\pm$ 0.05	 & 61.67 $\pm$ 5.07	 & 2.97 $\pm$ 0.11		\\
LSGAN	 & 33.38 $\pm$ 5.10	 & 2.28 $\pm$ 0.06	 & 75.88 $\pm$ 9.56	 & 3.02 $\pm$ 0.09		\\
WGAN-gp	 & 42.16 $\pm$ 2.77	 & 2.45 $\pm$ 0.07	 & 102.61 $\pm$ 18.48 & 2.90 $\pm$ 0.14		\\
Dragan	 & 20.36 $\pm$ 1.25	 & 2.35 $\pm$ 0.05	 & \textbf{50.41} $\pm$ 2.68	 & 2.91 $\pm$ 0.04		\\
SS-GAN	 & 21.49 $\pm$ 1.49	 & 2.37 $\pm$ 0.05	 & 61.20 $\pm$ 3.53	 & 2.97 $\pm$ 0.08		\\
NS-GAN	 & 20.50 $\pm$ 1.14	 & 2.37 $\pm$ 0.04	 & 60.02 $\pm$ 4.47	 & 2.98 $\pm$ 0.08		\\
NSS-GAN	 & \textbf{19.45} $\pm$ 1.15	 & 2.42 $\pm$ 0.05	 & 58.86 $\pm$ 3.92	 & 2.97 $\pm$ 0.08		\\
\hline      
 & \multicolumn{2}{c|}{Conference}  & \multicolumn{2}{c}{Flowers}   \\
 & FID & IS & FID & IS  \\
  \hline
GAN	 & 77.96 $\pm$ 5.49	 & 4.19 $\pm$ 0.11	 & 97.71 $\pm$ 4.91	 & 2.98 $\pm$ 0.09		\\
GAN-gr	 & 70.82 $\pm$ 3.80	 & 4.08 $\pm$ 0.11	 & 99.25 $\pm$ 5.59	 & 3.15 $\pm$ 0.09		\\
LSGAN	 & 81.90 $\pm$ 9.49	 & 4.01 $\pm$ 0.11	 & 121.81 $\pm$ 8.50	 & 2.72 $\pm$ 0.15		\\
WGAN-gp	 & 121.05 $\pm$ 7.60	 & 3.51 $\pm$ 0.13	 & 129.05 $\pm$ 7.70	 & 2.89 $\pm$ 0.08		\\
Dragan	 & 67.07 $\pm$ 5.45	 & 4.18 $\pm$ 0.16	 & 98.37 $\pm$ 6.63	 & 3.00 $\pm$ 0.09		\\
SS-GAN	 & 80.07 $\pm$ 5.44	 & 4.02 $\pm$ 0.10	 & 91.93 $\pm$ 5.12	 & 3.11 $\pm$ 0.09		\\
NS-GAN	 & 73.11 $\pm$ 4.07	 & 4.02 $\pm$ 0.08	 & 91.44 $\pm$ 4.47	 & 3.07 $\pm$ 0.08		\\
NSS-GAN	 & \textbf{64.61} $\pm$ 3.20	 & 4.21 $\pm$ 0.14	 & \textbf{86.28} $\pm$ 6.57	 & 3.09 $\pm$ 0.10		\\
\hline      
\end{tabular}      
\renewcommand{\arraystretch}{1}
\vspace{12pt} 
\end{table}         
%
%
%
%
%
%
\subsection{Comparison to state-of-the-arts}
We now compare the proposed NSS-GAN with the state-of-the-arts of GAN-optimization:
the baseline DCGAN (GAN),
GAN with the gradient regularization~\cite{Mescheder2018which},
LSGAN~\cite{mao2017LSGAN},
WGAN with gradient penalty (WGAN-gp)~\cite{Sailmans16improved},
Dragan~\cite{kodali2017convergence}, 
GANs using the scale-space (SS-GAN),
and the noise-space (NS-GAN) using the four of datasets.
Note that these GANs share the same backbone architecture of DCGAN.
In order to make our result independent to the hyper parameters of Adam, we employ grid search and tuned the learning-rate scale ($\eta$) in combination with the first momentum coefficient ($b_1$) for each pair of model and dataset, while the second momentum of Adam was fixed as $b_2=0.999$ based on our pretest.
%
%
For each model, we then choose the best condition using the mean FID within the 20 trials. Table~\ref{tab:tuned_condition} summarizes the tuned parameters in which the baseline GAN prefers a stable condition compared to the others.
\par
Figure~\ref{fig:final-comparison-images-part1} and Figure~\ref{fig:final-comparison-images-part2} present fake images by the tested GANs with their tuned hyper-parameters, where we use the generator of which FID is the closest to the mean of the independent trials.
Table~\ref{tab:final-comparison-metrics} summarizes the mean and std. of FID and IS of the experimented GANs for the four datasets within the 20 trials, where the proposed NSS-GAN with the constant parameter of $T=256$ has achieved better and comparable results than the state-of-the-arts, demonstrating the effectiveness of the presented NSS-GAN.
\par
More importantly, Table~\ref{tab:final-comparison-metrics} shows that the conventional scale-space (SS) GAN and the noise-space (NS) GAN were not consistently better than the baseline GAN.
In contrast, the proposed NSS-GAN has consistently outperformed the baseline GAN, SS-GAN, and NS-GAN in FID. This indicates that the stabilization effect of the presented NSS-GAN is not the simple summation of those by the scale-space with the noise-space but due to the better use of their mutually complementary relationship in the GAN optimization.
%
%
%
%
%
%
%
%
%

%
%
%
%
\begin{table}[htb]
\caption{The hyper-parameters of (left) StyleGAN2-Ada and (right)  StyleGAN2 with the proposed noisy scale-sapce: the R1 regularization ($\gamma$) of StyleGAN2 and the noise std of the proposed method ($\sigma$). We also employed the initial value $T=128$ for our noisy scale-space. }
\label{tab:tuned_condition:stylegan2}
%
\scriptsize
\centering
\renewcommand{\arraystretch}{0.8} 
\begin{tabular}{l c c c}
\hline
 &  & StyleGAN2-Ada & StyleGAN2-NSS \\
\hline
\multirow{2}{*}{MetFaces-$512$} & $\gamma$ & 1.64 & 0.82 \\
& $\sigma$ & - & 0.05 \\
\hline
\multirow{2}{*}{FFHQ-$512$} & $\gamma$ & 1.64 & 1.64 \\
& $\sigma$ & - & 0.1 \\
\hline
\end{tabular}
%
%
\end{table}
%
%
%

%
%
%
\subsection{Comparison to StyleGAN2-Ada}
As an additional study, we apply the noise scale-space to StyleGAN2-Ada~\cite{karras2020training} and compare it with the original algorithm using MetFaces dataset~\cite{karras2020training} and Flickr-Faces-HQ (FFHQ) dataset~\cite{karras2019style}.
StyleGAN2-Ada~\cite{karras2020training} is a state-of-the-art generative model for high-resolution images using a variety of data augmentations.  We denote our implementation by StyleGAN2-NSS. 
\par
StyleGAN2 series require 25Kimg-iterations to converge with images of the size $1024 \times 1024$. However, this requires a large-scale GPUs with a huge computational time. 
For the sake of reproducibility, we use down-scaled images of the size $512 \times 512$
that we call MetFaces-$512$ and FFHQ-$512$, respectively, and we also limit the training iterations to 5Kimg that is known to achieve reasonable results~\cite{karras2020training}. 
\par
For FFHQ-$512$ dataset, we implement our StyleGAN2-NSS by replacing the augmentation term (i.e., Ada)~\cite{karras2020training} by the proposed NSS function.
However, we have observed that both the vanilla StyleGAN2 without Ada and StyleGAN2-NSS can explode when applied to MetFaces-$512$. This means that the backbone network of StyleGAN2 does not work with MetFaces-$512$ without the Ada term. Thus, our implementation for MetFaces-$512$ includes the Ada term in combination with our NSS function.
\par
We follow the experimental set-up, the evaluation metrics, and the hyper-parameters of the official StyleGAN2 implementation, including the R1 regularization weight ($\gamma$) that we have tuned within $\gamma=0.82, 1.64, 3.28$ as recommended in~\cite{karras2020training}.
We tune the noise std of NSS based on the image variance curve of the images and also employ $T=128$ as the initial condition.
Table~\ref{tab:tuned_condition:stylegan2} summarizes the tuned hyper-parameters.
%
%
%
%
\par
Table~\ref{tab:result:stylegan2} summarizes FID and IS by StyleGAN2-Ada and StyleGAN2 using the proposed noisy scale-space.
Figure~\ref{fig:result:stylegan2} visualizes fake images created by the generator with the average FID within the individual trials.
Table~\ref{tab:result:stylegan2} and Figure~\ref{fig:result:stylegan2} show that the presented NSS-GAN has successfully improved the accuracy and quality of generated images using the state-of-art StyleGAN2.
\par
It has been reported\cite{karras2020training} that StyleGAN2-Ada outperforms Progressive-GAN~\cite{karras2017progressive} in accuracy. Therefore, our experimental results implicitly demonstrate that StyleGAN2-NSS is superior to Progressive-GAN. Moreover, our noisy scale-space is independent of the architecture in contrast to Progressive-GAN that requires the hierarchical architecture of networks.
%
%
%

%
%
%
%
\def \fw {165pt}
\begin{figure} [h!]
\small
\centering
\begin{tabular}{cc}
\includegraphics[width=\fw]{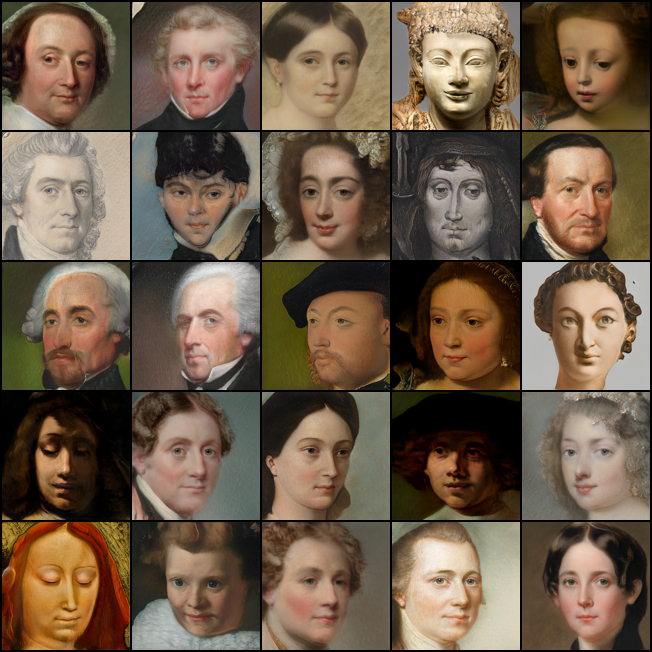} &
\includegraphics[width=\fw]{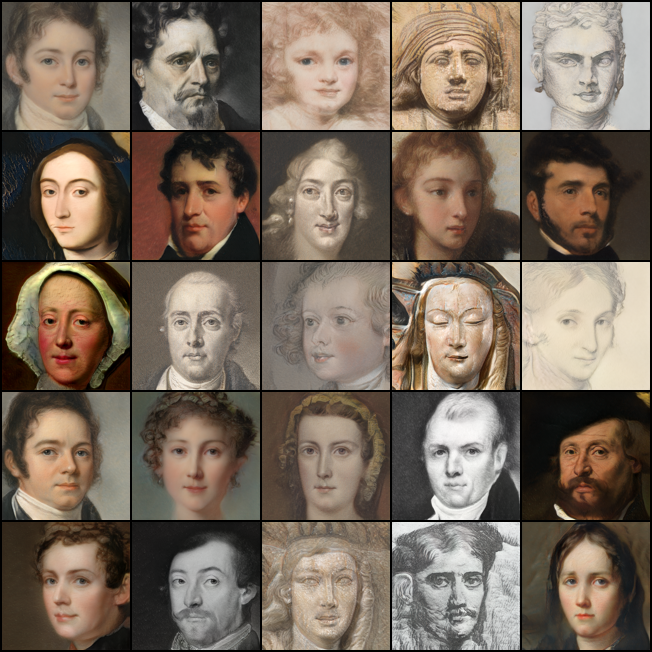} \\
\includegraphics[width=\fw]{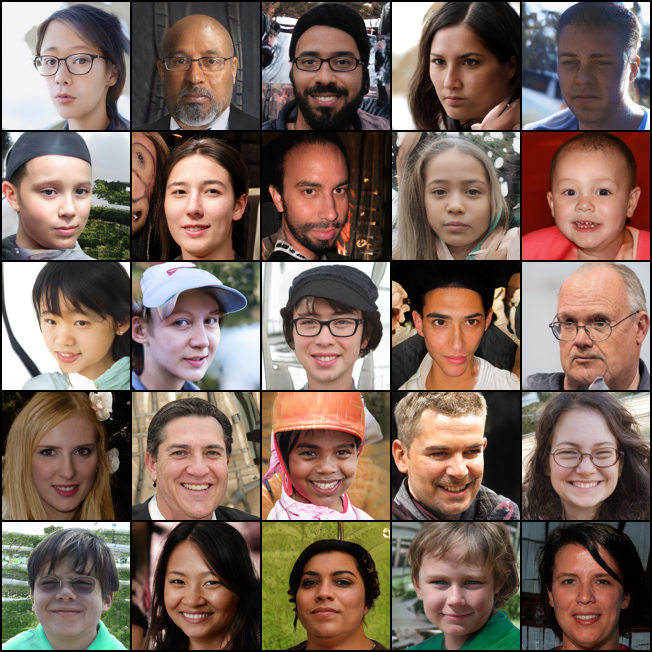} &
\includegraphics[width=\fw]{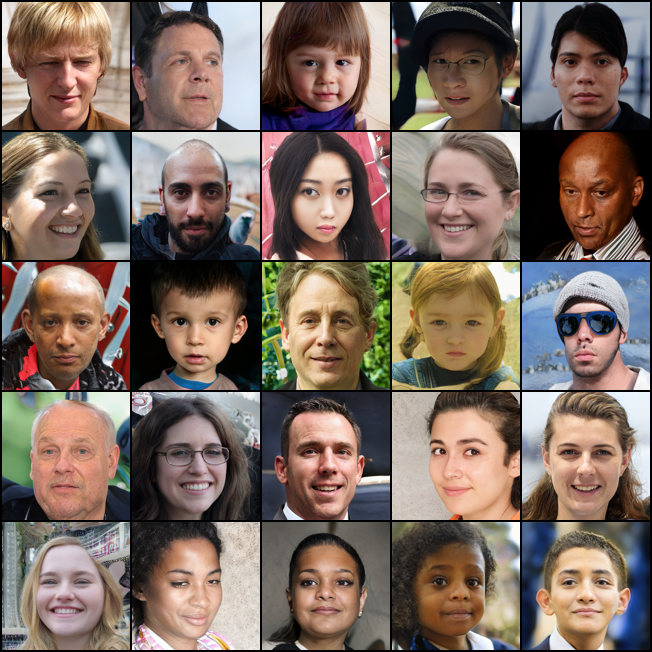} \\
StyleGAN2-Ada &
StyleGAN2-NSS (ours) \\
\end{tabular}
\caption{Fake images based on (top) MetFaces-$512$ and (bottom) FFHQ-$512$ datasets 
by (left) StyleGAN2-Ada and (right) StyleGAN2 with the proposed noisy scale-sapce trained  5KImg-iterations. The generators with the average FID within the 5 trials were used.}
\label{fig:result:stylegan2}
\end{figure}
%
%
%

%
%
%
\def \pw {100pt}
\begin{table} [htb]
\caption{Fr\'echet inception distance (FID-50K) and inception score (IS-50K) for (top part)  MetFaces-$512$ and (bottm part) FFHQ-$512$ datasets by (left) StyleGAN2-Ada and (right) StyleGAN2 with the proposed noisy scale-space. The models were trained 5KImg-iterations. The mean and std of the metrics were computed within the individual 5 trials.}
\label{tab:result:stylegan2}
\small
\centering
\begin{tabular}{l  c | P{\pw} P{\pw}}
\hline
 &  & StyleGAN2-Ada & StyleGAN2-NSS (ours) \\
\hline
\multirow{2}{*}{MetFaces-$512$} & FID ($\downarrow$) & 18.30 $\pm$ 1.87 & \textbf{17.25} $\pm$ 0.56 \\
& IS ($\uparrow$)    & 3.79 $\pm$ 0.14 & \textbf{3.87} $\pm$ 0.05 \\
\hline
\multirow{2}{*}{FFHQ-$512$} & FID ($\downarrow$) & 7.27 $\pm$ 0.25 & \textbf{5.52} $\pm$ 0.12 \\
& IS ($\uparrow$)    & 4.61 $\pm$ 0.03 & \textbf{ 4.91} $\pm$ 0.08 \\
\hline
\end{tabular}
\end{table}
%
%

%
%
%
%
\section{Conclusion} \label{sec:conclusion}
In the consideration of data manipulation for the stable optimization in GANs, we have proposed a discrete representation of data, called noisy scale-space (NSS), that gradually removes high-frequency information in image while adding noise, leading to a coarse-to-fine training of GANs.
In order to observe the side-effect of the conventional scale-space in GAN optimization, we have  proposed the synthetic dataset based on the Hadamard bases that visualizes the true distribution of the real and fake data.
We have experimented with the proposed NSS using two backbone networks: DCGAN and StyleGAN2 based on the major datasets for natural image generation tasks.
The experimental results have successfully demonstrated that: NSS-based GANs overtook the potential competitors and the state-of-the-arts in most cases.
\par
A limitation of our method is that we assume the diffusion can simplify (the real) data. Concretely, our NSS-GAN is inferior to the original GAN when using MNIST~\cite{lecun1998gradient} images that consist of 0/1 binary values. Obviously, smoothing the binary data increases the diversity of pixel values. Our assumption holds for natural images and our method yields the sufficient stabilization effect for GAN optimization irrespective of the content of images.
%
%
%
%
%
%

\bibliography{mybib}
\end{document}